\documentclass[10pt,twocolumn,letterpaper]{article}

\usepackage{cvpr}
\usepackage{times}
\usepackage{epsfig}
\usepackage{graphicx}
\usepackage{amsmath}
\usepackage{amssymb}
\usepackage{listings}
\usepackage{makecell}
\usepackage{booktabs}
\usepackage{multirow}
\usepackage{color}
\usepackage{comment}

\usepackage{icomma}
\usepackage{enumitem}
\usepackage{amsthm}
\setitemize{noitemsep,topsep=0pt,parsep=0pt,partopsep=0pt}
\setenumerate{noitemsep,topsep=0pt,parsep=0pt,partopsep=0pt}
\usepackage[bf,font=footnotesize]{caption}
\usepackage{subfig}
\usepackage[table]{xcolor} % Load last

\newcommand{\bx}{\mathbf x}

\newcommand{\bh}{\mathbf h}

\newcommand{\bv}{\mathbf v}

\newcommand{\bp}{\mathbf p}
\newcommand{\bw}{\mathbf w}
\newcommand{\bq}{\mathbf q}
\newcommand{\bg}{\mathbf g}
\newcommand{\bbf}{\mathbf f}

\newcommand{\bF}{\mathbf F}
\newcommand{\bX}{\mathbf X}

\newcommand{\bW}{\mathbf W}

\newcommand{\bH}{\mathbf H}

\newcommand{\bQ}{\mathbf Q}
\newcommand{\bV}{\mathbf V}

\newcommand{\helpers}{off-the-shelf CV methods\xspace}
\newcommand{\facts}{facts\xspace}

\newcommand{\CapFacts}{Facts\xspace}
\newcommand{\fact}{fact\xspace}
\newcommand{\sexyname}{VQA-Machine\xspace}

\def\T{{\!\top}}

\usepackage[pagebackref=true,breaklinks=true,letterpaper=true,colorlinks,bookmarks=false]{hyperref}

\cvprfinalcopy % *** Uncomment this line for the final submission

\newlength{\Oldarrayrulewidth}
\newcommand{\Cline}[2]{%
  \noalign{\global\setlength{\Oldarrayrulewidth}{\arrayrulewidth}}%
  \noalign{\global\setlength{\arrayrulewidth}{#1}}\cline{#2}%
  \noalign{\global\setlength{\arrayrulewidth}{\Oldarrayrulewidth}}}
\newcommand{\tableincell}[2]{\begin{tabular}{@{}#1@{}}#2\end{tabular}}

\ifcvprfinal\fi
\begin{document}

%%%%%%%%% TITLE
\title{The VQA-Machine: Learning How to Use Existing Vision Algorithms \\ to Answer New Questions}

\author{Peng Wang$^*$, Qi Wu\thanks{indicates equal contribution.}, Chunhua Shen, Anton van den Hengel\\
School of Computer Science, 
The University of Adelaide, Australia\\
{\tt\small \{p.wang,qi.wu01,chunhua.shen,anton.vandenhengel\}@adelaide.edu.au}}

\maketitle
%\thispagestyle{empty}

%%%%%%%%% ABSTRACT
\begin{abstract}
One of the most intriguing features of the Visual Question Answering (VQA) challenge is the unpredictability of the questions.   
Extracting the information required to answer them demands a variety of image operations from detection and counting, to segmentation and reconstruction.  
To train a method to perform even one of these operations accurately from  \{image,question,answer\} tuples would be challenging, but to aim to achieve them all with a limited set of such training data seems ambitious at best.  
We propose here instead a more general and scalable approach which exploits the fact that very good methods to achieve these operations already exist, and thus do not need to be trained.  
Our method thus learns how to exploit a set of external off-the-shelf algorithms to achieve its goal, an approach that has something in common with the Neural Turing Machine~\cite{graves2014neural}.
The core of our proposed method is a new co-attention model.
In addition, the proposed approach generates human-readable reasons for its decision, and can still be trained end-to-end without ground truth reasons being given. We demonstrate the effectiveness on two publicly available datasets, Visual Genome and VQA, and show that it produces the state-of-the-art results in both cases.
\end{abstract}

%%%%%%%%% BODY TEXT
\vspace{-10pt}
\section{Introduction}
\label{intro}

\begin{figure}[t!]
\centering
  \includegraphics[height=7.5cm,width=7.8cm]{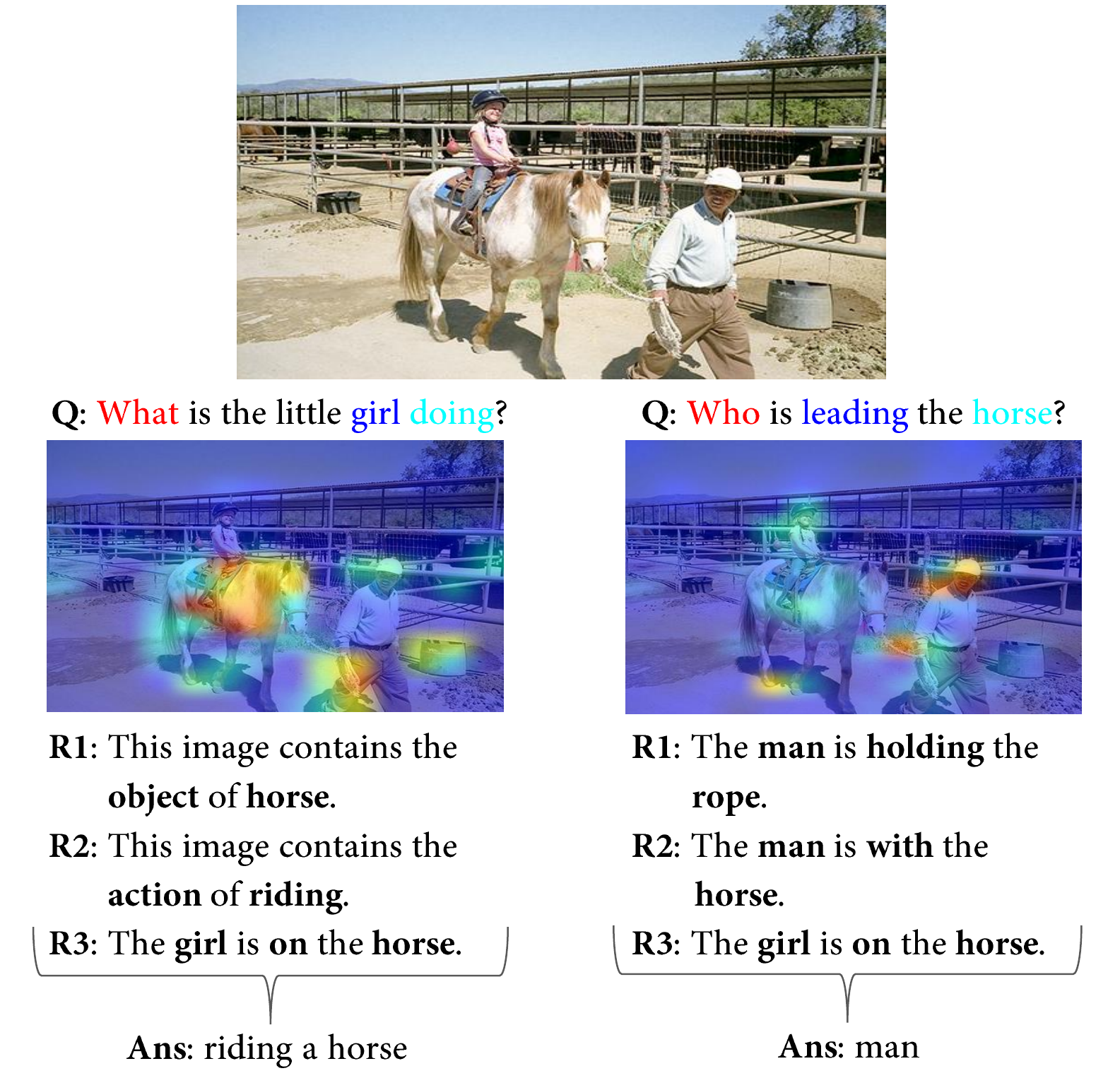}
  \vspace{-2pt}
  \caption{Two real example results of our proposed model. Given an image-question pair, our model  
generates not only an answer, but also a set of reasons (as text) and visual attention maps.
The colored words in the question have Top-3 weights, ordered as {\color{red}red}, {\color{blue}blue} and {\color{cyan}cyan}. The highlighted area in the attention map indicates the attention weights on the image regions. The Top-3 weighted visual \facts are re-formulated as human readable reasons. The comparison between the results for different questions relating to the same image shows that our model can produce highly informative reasons relating to the specifics of each question.}
  \label{img:example}
  \vspace{-18pt}
\end{figure}

Visual Question Answering (VQA) is an AI-complete task lying at the intersection of computer vision (CV) and natural language processing (NLP). 
Current VQA approaches are predominantly based on a joint embedding~\cite{antol2015vqa,gao2015you,malinowski2015ask,ren2015image,wu2015ask,zhou2015simple} of image features and question representations 
into the same space, the result of which is used to predict the answer.
One of the advantages of this approach is its ability to exploit a
pre-trained CNN model.

In contrast to this joint embedding approach we propose a co-attention based method which 
learns how to use a set of \helpers in answering image-based questions.
Applying the existing CV methods to the image generates a variety of information which we label the image \facts.  
Inevitably, much of this information would not be relevant to the particular question asked. So part of the role of the attention mechanism is to determine which types of \facts are useful in answering a question.

The fact that the \sexyname is able to exploit a set of \helpers in answering a question means that it does not need to learn how to perform these functions itself.  The method instead learns to predict the appropriate combination of algorithms to exploit in response to a previously unseen question and image.  It thus represents a step towards a Neural Network capable of learning an \emph{algorithm} for solving its problem.  In this sense it is comparable to the Neural Turning Machine~\cite{graves2014neural} (NTM) whereby an RNN is trained to use an associative memory module in solving its larger task.  The method that we propose does not alter the parameters of the external modules it uses, but then it is able to exploit a much wider variety of module types.

In order to enable the \sexyname to exploit a wide variety of available CV methods, and to provide a compact, but flexible interface, we formulate the visual \facts as triplets.
The advantages of this approach are threefold. Firstly, many relevant \helpers produce outputs that can be reformulated as triplets (see Tab.\ref{tab:triplet}).
Secondly, such compact formats are human readable, and interpretable. This allows us to provide human readable reasons along with the answers. See Fig.~\ref{img:example} for an example. At last, the proposed triplet representation is 
 similar to the triplet representations used in some Knowledge Bases, 
such as $<\textit{cat, eat, fish}>$. 
The method might thus be extendable to accept information from these sources.

To select the \facts which are relevant in answering a specific question, we employ a co-attention mechanism.
This is achieved by extending the approach of Lu \etal~\cite{lu2016hierarchical}, which proposed a co-attention mechanism that jointly reasons about 
the image and question, to also reason over a set of \facts.
Specifically, we design a sequential co-attention mechanism (see Fig.\ref{fig:co-atten}) which aims to ensure that attention can be passed effectively between all three forms of data.  
The initial question representation (without attention) is thus first used to guide \facts weighting. The weighted \facts and the initial question representation are then combined to guide the image weighting. The weighted \facts and image regions are then jointly used to guide the question attention mechanism. 
All that remains is to run the \fact attention again, but informed by the question and image attention weights, as this completes the circle, and means that each attention process has access to the output of all others.
All of the weighted features are further fed into a multi-layer perceptron (MLP) to predict the answer.

One of the advantages of this approach is that the question, image, and \facts are interpreted together, which particularly means that information extracted from the image (and represented as \facts) can guide the question interpretation. A question such as 
`Who is looking at the man with a telescope?' means different things when combined with an image of a man holding a telescope, rather than an image of a man viewed through a telescope.

Our main contributions are as follows:
\begin{itemize}
 \item We propose a new VQA model which is able to learn to adaptively combine multiple \helpers to answer questions.
 \item To achieve that, we
extend the co-attention mechanism to a higher order which is able to jointly process questions, image, and \facts.
 \item 
The method that we propose generates not only an answer to the posed questions, but also a set of supporting information, including the visual (attention) reasoning and human-readable textual reasons. To the best of our knowledge, this is the first VQA model that is capable of outputting human-readable reasons on free-form open-ended visual questions.
 \item Finally, we evaluate our proposed model on two VQA datasets.
Our model achieves the state-of-art in both cases. A human agreement study is conducted to evaluate the reason generation ability of our model.
 
\end{itemize}

\section{Related Work}
\label{rel_work}
\begin{figure*}[t!]
	\centering
	\includegraphics[type=pdf,ext=.pdf,read=.pdf,width=0.96\textwidth]{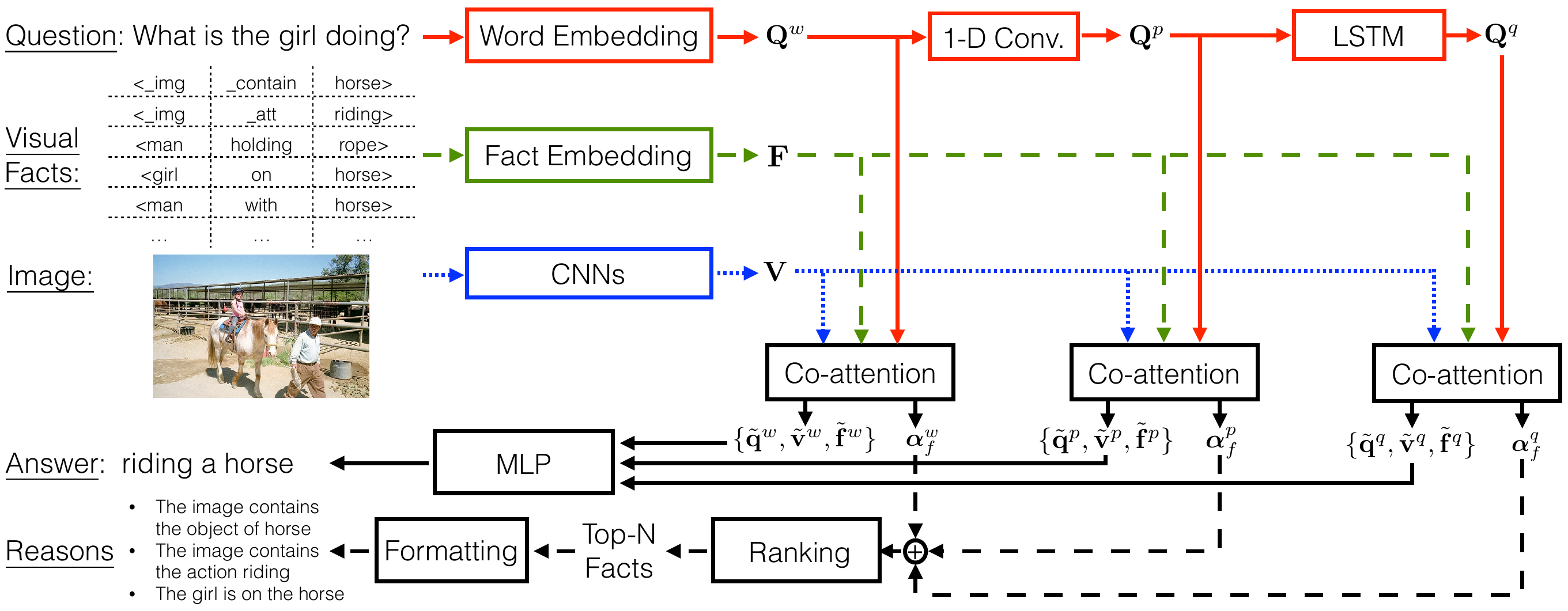}%
	\caption{The proposed VQA model. The input question, \facts and image features are weighted 
at three question-encoding levels. Given the co-weighted features at all levels, a multi-layer perceptron (MLP) classifier is used to predict answers. Then the ranked \facts are used to generate reasons.
	}
	\label{fig:system}
	\vspace{-10pt}
\end{figure*}

\vspace{-0.1cm}
\paragraph{Joint Embedding}
Most recent methods are based on a joint embedding of the image and the question using a deep neural network. Practically, image representations are obtained through CNNs that have been pre-trained on an object recognition task. The question is typically passed through an RNN, which produces a fixed-length vector representation. These two representations are jointly embedded into the same space and fed into a classifier which predicts the final answer. Many previous works~\cite{antol2015vqa,gao2015you,malinowski2015ask,ren2015image,wu2015image,zhou2015simple} adopted this approach, while some \cite{fukui2016multimodal,kim2016multimodal,saito2016dualnet} have proposed modifications of this basic idea. However, these modifications are either focused on developing more advanced embedding techniques
or employing different question encoding methods,
with only very few methods actually aim to improve the visual information available \cite{mallya2016learning,tommasi2016solving,wu2015ask}.
This seems a surprising situation given that VQA can be seen as encompassing the vast majority of CV tasks (by phrasing the task as a question, for instance).
It is hard to imagine that a single pre-trained CNN model (such as VGG~\cite{simonyan2014very} or ResNet~\cite{he2015resnet}) would suffice for all such tasks, or be able to recover all of the required visual information. 
The approach that we propose here, in contrast, is able to exploit the wealth of methods that already exist to extract useful information from images, and thus does not need to learn to perform these operations from a dataset that is ill suited to the task.  

\vspace{-0.4cm}
\paragraph{Attention Mechanisms}
Instead of directly using the holistic global-image embedding from the fully connected layer of a CNN, several recent works \cite{ilievski2016focused, lu2016hierarchical,shih2015look,xu2015ask,yang2015stacked,zhu2015visual7w} have explored image attention models for VQA. Specifically, the feature map (normally the convolutional layer of a pre-trained deep CNN) is used with the question to determine spatial weights that reflect the most relevant regions of the image. Recently, Lu \etal \cite{lu2016hierarchical} determine attention weights on both image regions and question words.
In our work, we extend the co-attention to a higher order so that the image, question and \facts can be jointly weighted.

\vspace{-0.4cm}
\paragraph{Modular Architecture and Memory Networks}
Neural Module Networks (NMNs) were introduced by Andreas \etal. in \cite{andreas2016learning,andreas2015deep}. 
In NMNs, the question parse tree is turned into an assembly of modules from a predefined set, which are then used to answer the question. 
Dynamic Memory Networks (DMN) \cite{xiong2016} retrieves the `facts' required to answer the question, where the `facts' are simply CNN features calculated over small image patches.
In comparison to NMNs and DMNs, our method uses a set of external algorithms that does not depend on the question.  The set of algorithms used is larger, however, the method for combining their outputs is more flexible, and varies in response to the question, the image, and the \facts. 

\vspace{-0.45cm}
\paragraph{Explicit Reasoning}
One of the limitations of most VQA methods is that it impossible to distinguish between an answer which has arisen as a result of the image content, and one selected because it occurs frequently in the training set~\cite{DBLP:journals/corr/TeneyLH16}. This is a significant limitation to the practical application of the technology, particularly in Medicine or Defence, as it makes it impossible to have any faith in the answers provided.
One solution to this problem is to provide human readable reasoning to justify or explain the answer, which has been a longstanding goal in Neural Networks (see~\cite{craven1994using,thrun1995extracting}, for example).
Wang \etal \cite{wang2015explicit} propose a VQA framework named ``Ahab'' that uses explicit reasoning over an RDF (Resource Description Framework) Knowledge Base to derive the answer, 
which naturally gives rise to a reasoning chain. This approach is limited to a hand-crafted set of question templates, however.
FVQA~\cite{wang2016fvqa} used an LSTM and a data-driven approach to learn the mapping of images/questions to RDF queries, but only considers questions relating to specified Knowledge Bases. In this work, we employ attention mechanisms over \facts provided by multiple \helpers. The \facts are formulated as human understandable structural triplets and are further processed into human readable reasons.
To the best of our knowledge, this is the first approach that can provide human readable reasons for the open-ended VQA problem. A human agreement study is reported in Sec.~\ref{sec:human} which demonstrates the performance of this approach.

\section{Models}
\label{model}
In this section, we introduce the proposed VQA model that takes questions, images and \facts as inputs and outputs a predicted answer with ranked reasons. The overall framework is described in Sec.~\ref{sec:overall}, while Sec.~\ref{sec:coatten} demonstrates how the three types of input are jointly embedded using the proposed sequential co-attention model. Finally, the module used for generating answers and reasons is introduced in Sec.~\ref{sec:mlp}.  

\subsection{Overall Framework}
\label{sec:overall}
The entire model is shown in the Fig.~\ref{fig:system}.  The first step sees the input question encoded at three different levels. 
At each level, the question features are embedded jointly with images and \facts via the proposed sequential co-attention model. Finally, a multi-layer perceptron (MLP) is used to predict answers based on the outputs (\ie, the weighted question, image and \fact features) of the co-attention models at all levels. Reasons are generated by ranking and reformulating the weighted \facts.

\vspace{-0.55cm}
\paragraph{Hierarchical Question Encoding}
We apply a hierarchical question encoding \cite{lu2016hierarchical} to effectively capture the information from a question at multiple scales, \ie word, phase and sentence level.
Firstly, the one-hot vectors of question words $\bQ = [ \bq_1, \dots, \bq_T  ]$ are embedded individually to continuous vectors 
$\bQ^w = [ \bq^w_1, \dots, \bq^w_T  ]$.
Then 1-D convolutions with different filter sizes (unigram, bigram and trigram) are applied to the word-level embeddings $\bQ^w$, followed by a max-pooling over different filters at each word location, to form the phrase-level features $\bQ^p = [ \bq^p_1, \dots, \bq^p_T  ] $.
Finally, the phrase-level features are further encoded by an LSTM, resulting the question-level features $\bQ^q = [ \bq^q_1, \dots, \bq^q_T  ]$.

\begin{table}[t!]
%\vspace{-5pt}
	\begin{center}
		\scriptsize
		\resizebox{\linewidth}{!}{
			\rowcolors{2}{gray!25}{} % Zebra rows (starting on row 2)
			\begin{tabular}{l l }
				\Xhline{2\arrayrulewidth}
									Triplet													&Example \\ \hline
				{\ttfamily (\_img,}{\ttfamily \_scene,}{\ttfamily img\_scn)}		& {\ttfamily (\_img,}{\ttfamily \_scene,}{\ttfamily office)}		 \\
				{\ttfamily (\_img,}{\ttfamily \_att,}{\ttfamily img\_att)}	& {\ttfamily (\_img,}{\ttfamily \_att,}{\ttfamily wedding)}		 \\
				{\ttfamily (\_img,}{\ttfamily \_contain,}{\ttfamily obj)}			& {\ttfamily (\_img,}{\ttfamily \_contain,}{\ttfamily dog)}	\\ 
				{\ttfamily (obj,}{\ttfamily \_att,}{\ttfamily obj\_att)}		& {\ttfamily (shirt,}{\ttfamily \_att,}{\ttfamily red)}												\\
				{\ttfamily (obj1,}{\ttfamily rel,}{\ttfamily obj2)}					& {\ttfamily (man,}{\ttfamily hold,}{\ttfamily umbrella)}	\\
				\Xhline{2\arrayrulewidth}
			\end{tabular}}
			\vspace{-1pt}
			\caption{\CapFacts represented by triplets. 
				{\ttfamily \_img}, {\ttfamily \_scene}, {\ttfamily \_att} and {\ttfamily \_contain} are specific tokens. 
				While {\ttfamily img\_scn}, {\ttfamily obj}, {\ttfamily img\_att}, {\ttfamily obj\_att} and {\ttfamily rel} refer to 
							vocabularies describing image scenes, objects, image/object attributes and relationships between objects.
				} 
			\label{tab:triplet}
			\vspace{-22pt}
		\end{center}
	\end{table}

\vspace{-0.5cm}
\paragraph{Encoding Image Regions}
Following \cite{lu2016hierarchical,yang2015stacked}, the input image is resized to $448 \times 448$ and divided to $14 \times 14$ regions.
The corresponding regions of the last pooling layer of VGG-$19$~\cite{simonyan2014very} or ResNet-$100$~\cite{simonyan2014very} networks are extracted and further embedded using a learnable embedding weight.
The outputs of the embedding layer, denoted as $\bV = [ \bv_1, \dots, \bv_{N} ]$ ($N = 196$ is the number of image regions), are taken as image features.

\vspace{-0.45cm}
\paragraph{Encoding \CapFacts}
In this work, we use triplets of the form
{\ttfamily (subject,}{\ttfamily relation,}{\ttfamily object)}
to represent \facts in an image, where {\ttfamily subject} and {\ttfamily object} denote two 
visual concepts and {\ttfamily relation} represents a relationship between these two concepts.
This format of triplets is very general and widely used in large-scale structured knowledge graphs (such as DBpedia~\cite{auer2007dbpedia}, Freebase~\cite{bollacker2008freebase}, YAGO~\cite{mahdisoltani2014yago3}) to
record a surprising variety of information.
In this work, we consider five types of visual concepts as shown in Table~\ref{tab:triplet},
which respectively records information about the image scene, objects in the image, attributes of objects and the whole image,
and relationships between two objects.
Vocabularies are constructed for {\ttfamily subject}, {\ttfamily relation} and {\ttfamily object},
and the entities in a triplet are represented by individual one-hot vectors.
Three embeddings are learned end-to-end  to project the three triplet entities to continuous-valued vectors respectively
($\bbf_{s}$ for {\ttfamily subject}, $\bbf_{r}$ for {\ttfamily relation}, $\bbf_{o}$ for {\ttfamily object}).
The concatenated vector $\bbf = [ \bbf_{s}; \bbf_{r}; \bbf_{o} ]$ is then used to represent the corresponding \fact. 
By applying different types of visual models, we achieve a list of encoded \fact features $\bF = [\bbf_1, \dots, \bbf_M ]$, where $M$ is the number of extracted \facts.
Note that this approach may be easily extended to using any existing vision methods to extract image information that might usefully be recorded as a triplet, or even more generally, to any method which generates output which can be encoded as a fixed-length vector.

\subsection{Sequential Co-attention}
\label{sec:coatten}

\begin{figure}
\vspace{-3pt}
	\centering
	\includegraphics[type=pdf,ext=.pdf,read=.pdf,width=6cm,height=6.5cm]{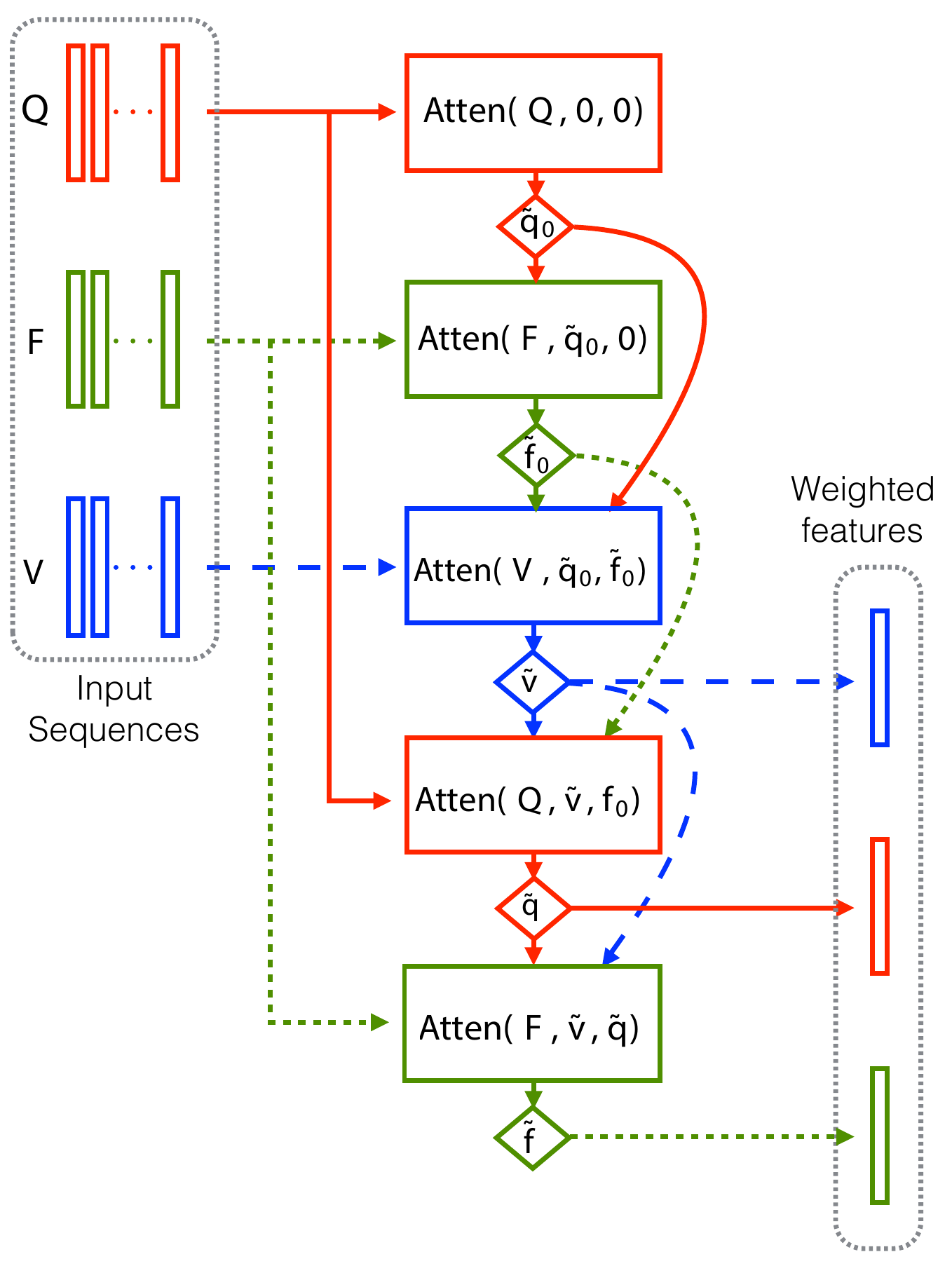}%
	\vspace{-10pt}
	\caption{The sequential co-attention module. Given the feature sequences for the question ($\bQ$), \facts ($\bF$) and 
		image ($\bV$), this module sequentially generates 
		weighted
		features ($\tilde{\bv}$, $\tilde{\bq}$, $\tilde{\bbf}$).
	}
	\vspace{-12pt}
	\label{fig:co-atten}
\end{figure}

Given the encoded question/image/\fact features, the proposed co-attention approach sequentially generates attention weights for each feature type using the other two as guidance, as shown in Fig~\ref{fig:co-atten}.
The operation in each of the five attention modules is denoted $\tilde{\bx} = \mathrm{Atten}(\bX, \bg_1, \bg_2)$,
which can be expressed as follows:
\vspace{-2pt}
\begin{subequations}
\label{eq:seq_attend}
\begin{align}
\bH_i &= \mathrm{tanh} ( \bW_x \bx_i \!+\! \bW_{g_1} \bg_1 \!+\! \bW_{g_2} \bg_2 ),  \\
\alpha_i &= \mathrm{softmax}(\bw^\T \bH_i), \quad i = 1,\dots,N, \\
\tilde{\bx}\,\, &= \textstyle{\sum_{i=1}^{N}} \alpha_i \bx_i,
\end{align}
\end{subequations}
where $\bX = [ \bx_i, \dots, \bx_N ] \in \mathcal{R}^{d \times N}$ is the input sequence, 
and the fixed-length vectors $\bg_1$, $\bg_2 \in \mathcal{R}^d$ are attention guidances.
$\bW_x$, $\bW_{g_1}$, $\bW_{g_2} \in \mathcal{R}^{h \times d}$ and $\bw \in \mathcal{R}^h$ are embedding parameters 
to be learned. Here
${\boldsymbol \alpha}$ is the attention weights of the input sequence features and the weighted sum $\tilde{\bx}$ is the weighted feature.

In the proposed co-attention approach, the encoded question/image/fact features (see Sec.~\ref{sec:overall}) are sequentially fed into the attention module (Equ.\ref{eq:seq_attend}) as input sequences, and the weighted features from the previous two steps are
used as guidance. 
Firstly, the question features are summarized without any guidance ($\tilde{\bq}_0 = \mathrm{Atten}(\bQ, \mathbf{0}, \mathbf{0})$). 
At the second step, the \fact features are weighted based on the summarized question features ($\tilde{\bbf}_0 = \mathrm{Atten}(\bF, \tilde{\bq}_0, \mathbf{0})$).
Next, the weighted image features are generated with the weighted \fact features and summarized question features
as guidances 
($\tilde{\bv} = \mathrm{Atten}(\bV, \tilde{\bq}_0, \tilde{\bbf}_0)$).
In step 4 
($\tilde{\bq} = \mathrm{Atten}(\bQ, \tilde{\bv}, \tilde{\bbf}_0)$)
and step 5  
($\tilde{\bbf} = \mathrm{Atten}(\bF, \tilde{\bv}, \tilde{\bq})$),
the question and \fact features are re-weighted based on the outputs of the previous steps.
Finally, the weighted question/image/\fact features ($\tilde{\bq}$, $\tilde{\bbf}$, $\tilde{\bv}$) are further used for answer prediction and the attention weights of the last attention module ${\boldsymbol \alpha}_f$ are used for reasons generation.

\vspace{-2pt}
\subsection{Answer Prediction and Reason Generation}
\label{sec:mlp}
\vspace{-3pt}

Similar to many previous VQA models \cite{lu2016hierarchical,Noh2015image,ren2015image,zhou2015simple}, the answer prediction process is treated as a multi-class classification problem, in which each class corresponds to a distinct answer. Given the weighted features generated from the word/phrase/question levels, a multi-layer perceptron (MLP) is used for classification:
\vspace{-2pt}
\begin{subequations}
	\label{eq:mlp}
	\begin{eqnarray}
	&\bh^w &= \mathrm{tanh} \big( \bW_w ( \tilde{\bq}^w + \tilde{\bv}^w + \tilde{\bbf}^w) \big), \\
	&\bh^p &= \mathrm{tanh} \big( \bW_p \big[ (\tilde{\bq}^p + \tilde{\bv}^p + \tilde{\bbf}^p); \bh^w \big] \big), \\
	&\bh^q &= \mathrm{tanh} \big( \bW_q \big[ (\tilde{\bq}^q + \tilde{\bv}^q + \tilde{\bbf}^q); \bh^p \big] \big), \\
	&\bp   &= \mathrm{softmax} (\bW_h \bh^q),
	\end{eqnarray}
\end{subequations}
where 
$\{ \tilde{\bq}^w, \tilde{\bv}^w, \tilde{\bbf}^w \}$,
$\{ \tilde{\bq}^p, \tilde{\bv}^p, \tilde{\bbf}^p \}$ and
$\{ \tilde{\bq}^q, \tilde{\bv}^q, \tilde{\bbf}^q \}$
are weighted features from all three levels.
$\bW_w$, $\bW_p$, $\bW_q$ and $\bW_h$ are parameters and $\bp$ is the probability vector. 

As shown in Fig.~\ref{fig:system}, the attention weights of \facts from all three levels 
(${\boldsymbol \alpha}_f^w$, ${\boldsymbol \alpha}_f^p$, ${\boldsymbol \alpha}_f^q$) are 
summed together. Then the top-$3$ ranked \facts are automatically formulated (by simple rule-based approaches, see supplementary) to human readable sentences and are considered as reasons.

\vspace{-0.1cm}
\section{Experiments}
\label{exp}

We evaluate our models on two datasets, Visual Genome QA \cite{krishnavisualgenome} and VQA-real \cite{antol2015vqa}. The Visual Genome QA 
contains 1,445,322 questions on 108,077 images. 
Since the official split of the Visual Genome dataset is not released, we randomly generate our own split. In this split, we have 723,060 training, 54,506 validation and 667,753 testing question/answer examples, based on 54,038 training, 4,039 validation and 50,000 testing images. VQA dataset \cite{antol2015vqa} is one of the most widely used datasets, which comprises two parts, one using natural images, and a second using cartoon images. In this paper, we only evaluate our models on the real image subset, which we have labeled VQA-real. VQA-real comprises 123,287 training and 81,434 test images. 

\subsection{Implementation Details}
\label{imp_details}

\vspace{-0.1cm}
\paragraph{\CapFacts Extraction}
Using the training splits of the Visual Genome dataset, we trained three naive multi-label CNN
models to extract objects, object attributes and object-object relations.
Specifically, we extract the top-$5000$/$10000$/$15000$ triplets of the form {\ttfamily (\_img,}{\ttfamily \_contain,}{\ttfamily obj)}/{\ttfamily (obj,}{\ttfamily \_att,}{\ttfamily obj\_att)}/
{\ttfamily (obj1,}{\ttfamily rel,}{\ttfamily obj2)} based on the annotations of Visual Genome and use them as class labels. We then formulate them as three separate multi-label classification problems using an element-wise logistic loss function. The pre-trained VGG-$16$ model is used as initialization and only the fully-connected layers are fine-tuned. We also used the scene classification model of \cite{zhou2014learning} and the image attribute model of \cite{wu2015image}
to extract \facts not included in the Visual Genome dataset (\ie, image scenes and image attributes). Thresholds are set for these visual models and on average $76$ \facts are extracted for each image. The confidence score of the predicted \facts is added as an additional dimension of encoded \fact features $\bbf$.

\vspace{-0.3cm}
\paragraph{Training Parameters}
In our system, the dimensions of the encoded question/image/\fact features ($d$ in Eq.~\ref{eq:seq_attend}) and the hidden layers of the LSTM and co-attention models ($h$ in Eq.~\ref{eq:seq_attend}) are set to $512$. 
For \facts, the {\ttfamily subject}/{\ttfamily relation}/{\ttfamily object} entities are embedded  to $128$/$128$/$256$ dimensional vectors respectively and concatenated to form $512$d vectors.
We used two layers of LSTM model.
For the MLP in Eq.~\eqref{eq:mlp}, the dimensions of $\bh^w$ and $\bh^p$ are also $512$, while the dimension of $\bh^q$ is 
set to $1024$ for the VQA dataset and $2048$ for the Visual Genome dataset.
For prediction, we take the top $3000$ answers for the VQA dataset and the top $5000$ answers for the Visual Genome dataset.
The whole system is implemented on the Torch7~\cite{collobert2011torch7}\footnote{The code and pre-trained models will be released upon the acceptance of the paper.} and trained end-to-end but with fixed CNN features. 
For optimzation, the \texttt{RMSProp} method is used with a base learning rate of $2\times10^{-4}$
and momentum $0.99$. The model is trained for up to $256$ epochs until the validation error has not improved in the last $5$ epochs.

\subsection{Results on the Visual Genome QA}
%\vspace{-0.2cm}
\paragraph{Metrics}
We use the accuracy value and the Wu-Palmer similarity (WUPS) \cite{wu1994verbs} to measure the performance on the Visual Genome QA (see Tab.\ref{vg_results}). 
Before comparison, all responses are made lowercase, numbers converted to digits, and punctuation~\&~articles removed. The accuracy according to the question types are also reported.

\vspace{-0.3cm}
\paragraph{Baselines and State-of-the-art}
The first baseline method is \textbf{VGG+LSTM} from Antol \etal in \cite{antol2015vqa}, who uses a two layer LSTM to encode the questions and the last hidden layer of VGG \cite{simonyan2014very} to encode the images. The image features are then $\ell2$ normalized. 
We use the author provided code\footnote{https://github.com/VT-vision-lab/VQA\_LSTM\_CNN} to train the model on the Visual Genome QA training split. 
\textbf{VGG+Obj+Att+Rel+Extra+LSTM} uses the same configuration as VGG+LSTM, except that we concatenate additional image features extracted by VGG-16 models (fc7) that have been pre-trained on different CV tasks described in the previous section. \textbf{HieCoAtt-VGG} is the original model presented in \cite{lu2016hierarchical}, which is the current state of art. The authors' implementation\footnote{https://github.com/jiasenlu/HieCoAttenVQA} is used to train the model.% on the Visual Genome QA training split. 

\begin{table*}[t]
\centering
\resizebox{\linewidth}{!}{
%\rowcolors{1}{gray!25}{} % Zebra rows (starting on row 2)
\begin{tabular}{lcccccccccc}
\Xhline{2\arrayrulewidth}
\multicolumn{1}{c}{}                                      & \multicolumn{7}{c}{Accuracy (\%)}                                      &&   \multicolumn{2}{c}{WUPS (\%)}    \\  \Cline{0.8pt}{2-8} \Cline{0.8pt}{10-11} 
\multicolumn{1}{c}{Methods}                               & What     & Where    & When    & Who     & Why     & How      & \multirow{2}{*}{\textbf{Overall}} && \multicolumn{2}{c}{Overall}\\
\multicolumn{1}{c}{}                                      & (60.5\%) & (17.0\%) & (3.5\%) & (5.5\%) & (2.7\%) & (10.8\%) &         &   &@0.9          &@0.0          \\ \Xhline{2\arrayrulewidth}
\rowcolor{gray!30}VGG+LSTM \cite{antol2015vqa}                               & 35.12    & 16.33    & 52.71   & 30.03   & 11.55   & 42.69    & 32.46   &   & 38.30             & 58.39             \\
VGG+Obj+Att+Rel+Extra+LSTM                                & 36.88    & 16.85    & 52.74   & 32.30   & 11.65   & 44.00    & 33.88   &   & 39.61             & 58.89             \\ 
\rowcolor{gray!30}HieCoAtt-VGG \cite{lu2016hierarchical}                               & 39.72    & 17.53    & 52.53   & 33.80   & 12.62   & 45.14    & 35.94   &   & 41.75             & 59.97             \\ \Xhline{1.5\arrayrulewidth}
Ours-GtFact(Obj)                                          & 37.82    & 17.73    & 51.48   & 37.32   & 12.84   & 43.10    & 34.77   &   & 40.83             & 59.69             \\
\rowcolor{gray!25}Ours-GtFact(Obj+Att)                    & 42.21    & 17.56    & 51.89   & 37.45   & 12.93   & 43.90    & 37.50   &   & 43.24             & 60.39             \\
Ours-GtFact(Obj+Rel)                                      & 38.25    & 18.10    & 51.13   & 38.22   & 12.86   & 43.32    & 35.15   &   & 41.25             & 59.91             \\
\rowcolor{gray!25}Ours-GtFact(Obj+Att+Rel)                & 42.86    & 18.22    & 51.06   & 38.26   & 13.02   & 44.26    & 38.06   &   & 43.86             & 60.72             \\
Ours-GtFact(Obj+Att+Rel)+VGG                              & 44.28    & 18.87    & 52.06   & 38.87   & 12.93   & 46.08    & 39.30   &   & 44.94             & 61.21             \\\Xhline{1.5\arrayrulewidth}
\rowcolor{gray!20}Ours-PredFact(Obj+Att+Rel)              & 37.13    & 16.99    & 51.70   & 33.87   & 12.73   & 42.87    &         34.01&  & 39.92             &     59.20         \\
Ours-PredFact(Obj+Att+Rel+Extra)                          & 38.52    & 17.86    & 51.55   & 34.65   & 12.87   & 44.34    & 35.20   &   & 41.08             & 59.75             \\
\rowcolor{gray!20}Ours-PredFact(Obj+Att+Rel)+VGG          & 40.34    & 17.80    & 52.12   & 34.98   & 12.78   & 45.37    &         36.44&   & 42.16             &    60.09          \\
Ours-PredFact(Obj+Att+Rel+Extra)+VGG                      & 40.91    & 18.33    & 52.33   & 35.50   & 12.88   & 46.04    & 36.99   &   & 42.73             & 60.39             \\\Xhline{2\arrayrulewidth}
%\rowcolor{gray!20}Ours-PredFact(Obj+Att+Rel+Extra)+ResNet &          &          &         &         &         &          &         &   &              &              \\ \Xhline{2\arrayrulewidth}
\end{tabular}}
\caption{Ablation study on the Visual Genome QA dataset. Accuracy for different question types are shown. The percentage of questions for each type is shown in parentheses. We additionally calculate the WUPS at 0.9 and 0.0 for different models.}
\vspace{-10pt}
\label{vg_results}
\end{table*}

\vspace{-0.3cm}
\subsubsection{Ablation Studies with Ground Truth \CapFacts}
\vspace{-0.1cm}
In this section, we conduct an ablation study to evaluate the effectiveness of incorporating different types of \facts. To avoid the bias that would be caused by the varying accuracy with which the \facts are predicted, we use the ground truth \facts provided by the Visual Genome dataset as the inputs to our proposed models for ablation testing.

\textbf{GtFact(Obj)} is the initial implementation of our proposed co-attention model, using only the `object' \facts, which means the co-attention mechanisms only occur between question and the input object \facts. The overall accuracy of this model on the test split is 34.77\%, which already outperforms the baseline model \textbf{VGG+LSTM}. However, there is still a gap between this model and the \textbf{HieCoAtt-VGG} (35.94\%), which applies the co-attention mechanism to the questions and whole image features. Considering that the image features are still not used at this stage, this result is reasonable. 

Based on this initial model, we add the `attribute' and `relationship' \facts separately, producing two models, \textbf{GtFact(Obj+Att)} and \textbf{GtFact(Obj+Rel)}. Table \ref{vg_results} shows that the former performs better than the latter (37.50\% VS. 35.15\%), although both outperform the previous `Object'-only model. This suggests that `attribute' \facts are more effective than the `relationship' \facts. However, when it comes to  questions starting with `where' and `who', the \textbf{GtFact(Obj+Rel)} performs slightly better (18.10\% VS. 17.56\%, 38.22\% VS.37.45\%), which suggests that the `relationship' \facts play a more important role in these types of question. This makes sense because `relationship' \facts naturally encode location and identity information, for example $<cat, on, mat>$ and $<boy, holding, ball>$. The \textbf{GtFact(Obj+Att)} model exceeds the image features based co-attention model \textbf{HieCoAtt-VGG} by a large margin, 1.56\%. This is mainly caused by the substantial improvement in the `what' questions, from 39.72\% to 42.21\%. This is not surprising because the `attributes' \facts include detailed information relevant to the `what' questions, such as `color', `shape', `size', `material' and so on.

In the \textbf{GtFact(Obj+Att+Rel)}, all of the \facts are plugged into the proposed model, which brings further improvements for nearly all question types, achieving an overall accuracy 38.06\%. Compared with the baseline model \textbf{VGG+Obj+Att+Rel+Extra+LSTM} that uses a conventional method (concatenating and embedding) to introduce additional features into the VQA, our model outperforms by 4.18\%, which is a large gap. However, for the `when' questions, we find that the performance of our `\facts' based models are always lower than the image-based ones. We observed that this mainly because the annotated \facts in the Visual Genome normally do not cover the `when' related information, such as time and day or night. The survey paper \cite{wu2016visual} makes a similar observation - \textit{``98.8\% of the `when' questions in the Visual Genome can not be directly answered from the provided scene graph annotations''}. Hence, in the final model \textbf{GtFact(Obj+Att+Rel)+VGG}, we add the image features back, allowing for a higher order co-attention between image, question and \facts, achieving an overall accuracy 39.30\%. It also brings 1\% improvements on the `when' questions. The WUPS evaluation exhibits the same trend as the above results, the question type-specific results can be found in the supplementary material.

\begin{figure*}[t]
\resizebox{\linewidth}{!}{
\begin{tabular}{p{4.9cm}p{4.5cm}p{4.6cm}p{4.5cm}p{4.5cm}}
\includegraphics[height=3cm]{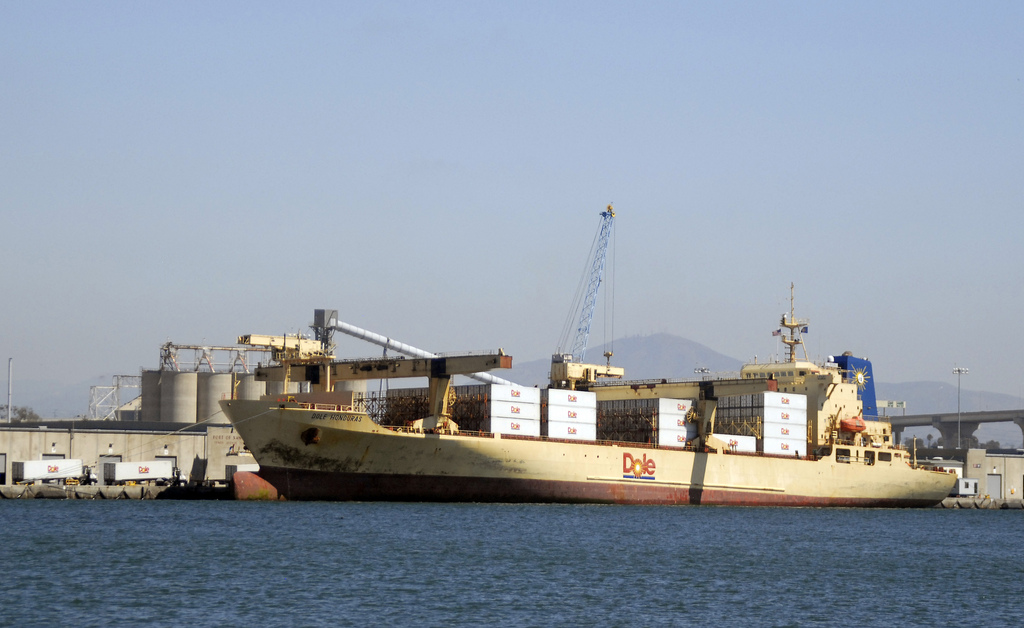}&\includegraphics[height=3cm]{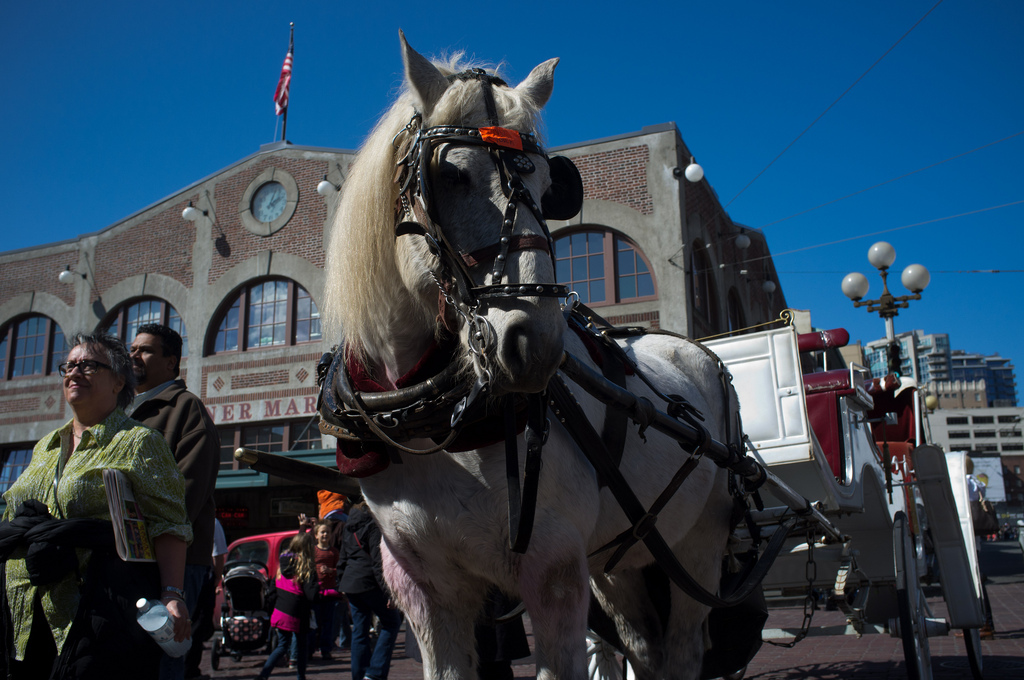} &\includegraphics[height=3cm]{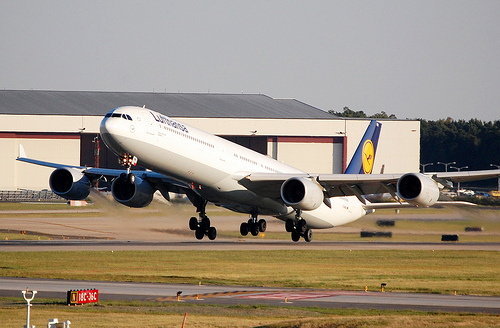} &\includegraphics[height=3cm,width=4.5cm]{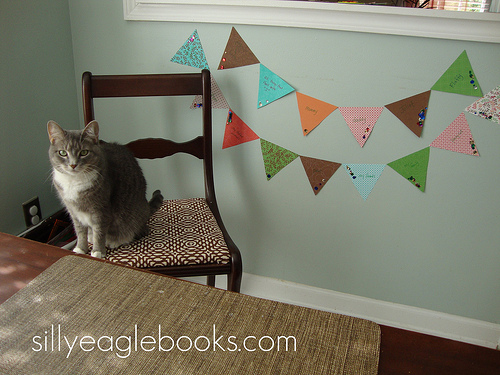} & \includegraphics[height=3cm]{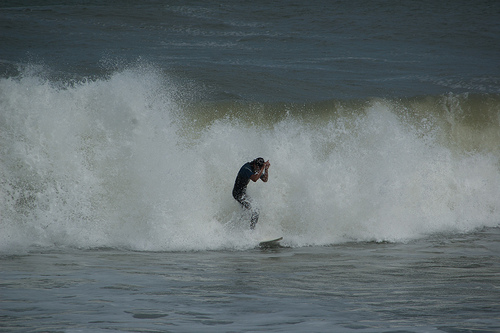} \\ 
\textbf{Q:} {\color{cyan}What} is in {\color{blue}the} {\color{red}water}?&\textbf{Q:} What time {\color{blue}of} {\color{red}day} {\color{cyan}is} it?&\textbf{Q:} What {\color{red}color} is the {\color{cyan}plane}'s {\color{blue}tail}?&\textbf{Q:} {\color{cyan}What} is sitting on {\color{red}the} {\color{blue}chair}?&\textbf{Q:} {\color{red}Who} {\color{cyan}is} {\color{blue}surfing}?\\
\textbf{A:} boat&\textbf{A:} daytime&\textbf{A:} blue&\textbf{A:} cat&\textbf{A:} man\\
%&&&&\\
\includegraphics[height=3cm]{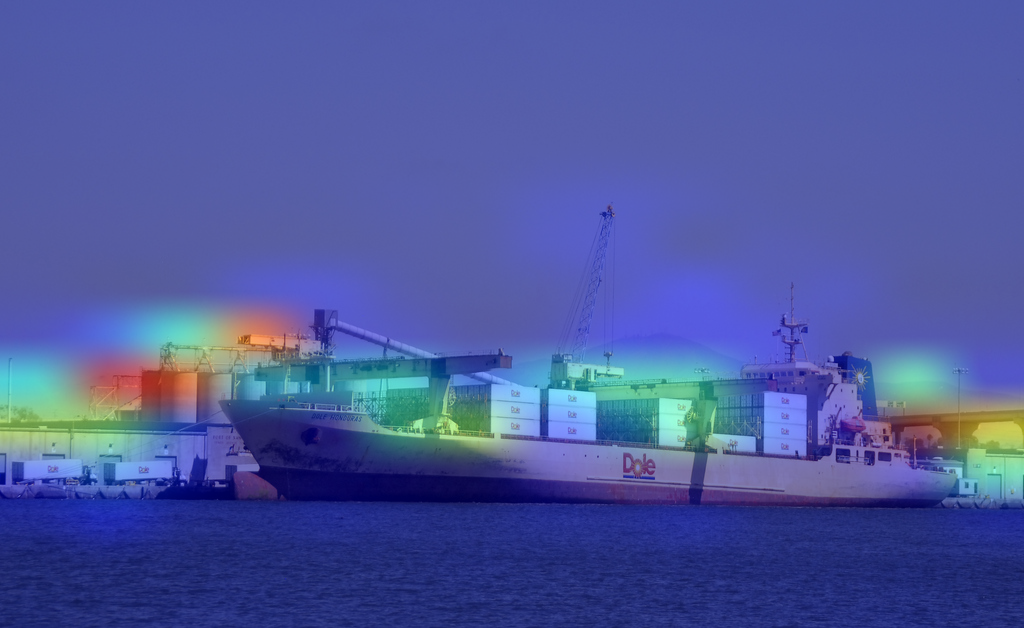}&\includegraphics[height=3cm]{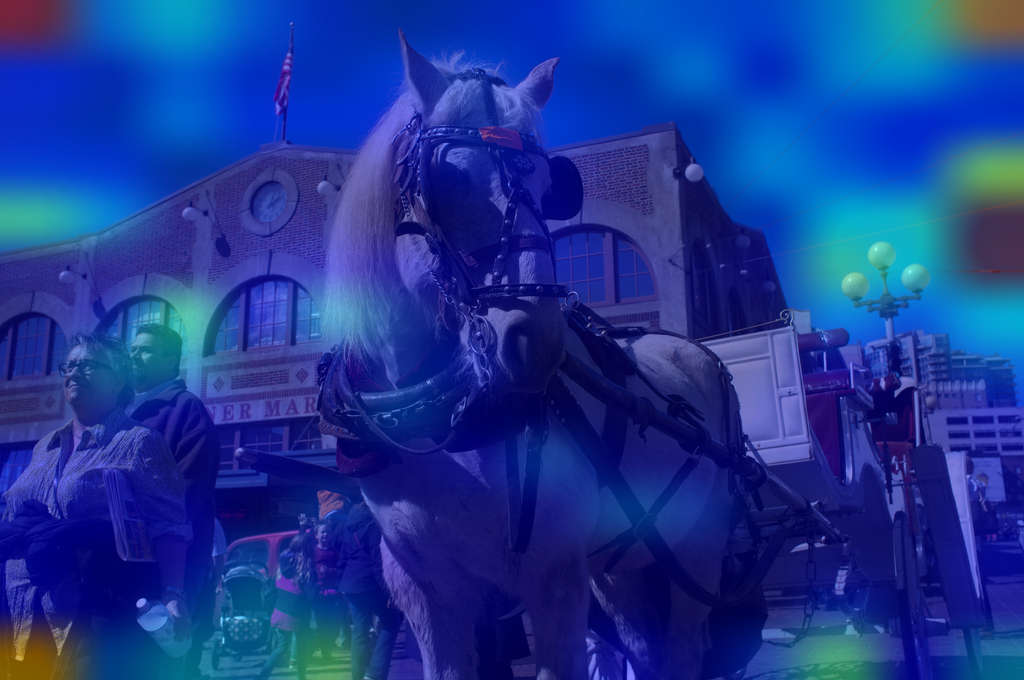} &\includegraphics[height=3cm]{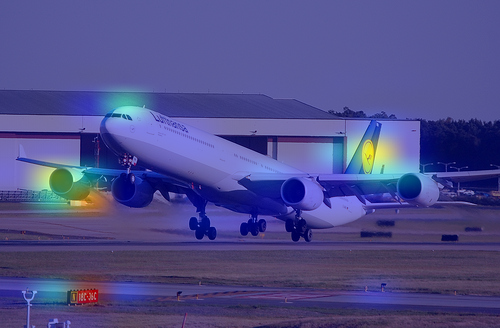} &\includegraphics[height=3cm,width=4.5cm]{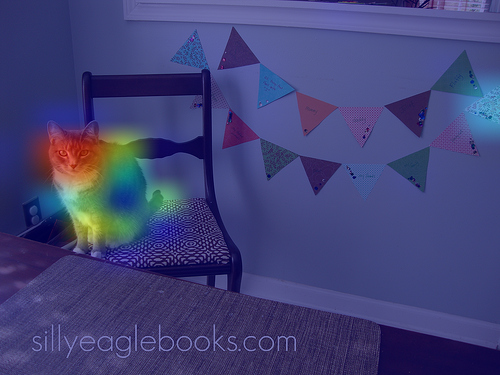} & \includegraphics[height=3cm]{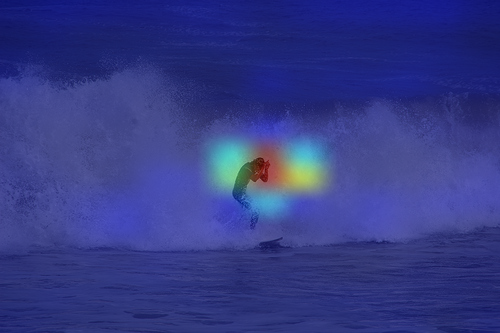} \\
%&&&&\\
\tableincell{p{4.8cm}}{$\bullet$ This image contains the \textbf{object} of \textbf{boat}. \\ $\bullet$ The \textbf{boat} is \textbf{on} the \textbf{water}.\\ $\bullet$ This image happens in the \textbf{scene} of \textbf{harbor}.} &\tableincell{p{4.5cm}}{$\bullet$ The \textbf{sky} is \textbf{blue}.\\ $\bullet$ The \textbf{horse} is \textbf{pulling}. \\ $\bullet$ This image contains the \textbf{object} of \textbf{road}.}&\tableincell{p{4.4cm}}{$\bullet$ The \textbf{tail} is \textbf{blue}.\\ $\bullet$ This image contains the \textbf{object} of \textbf{airplane}. \\ $\bullet$ This image happens in the \textbf{scene} of \textbf{airport}.}&\tableincell{p{4.4cm}}{$\bullet$ This image contains the \textbf{object} of \textbf{cat}.\\ $\bullet$ The \textbf{cat} is \textbf{on} the \textbf{chair}. \\ $\bullet$ The \textbf{fur} is \textbf{on} the \textbf{cat}}&\tableincell{p{4.4cm}}{$\bullet$ The \textbf{man} is \textbf{riding} the \textbf{surfboard}.\\ $\bullet$ The \textbf{man} is \textbf{wearing} the \textbf{wetsuit}. \\ $\bullet$ This image contains the \textbf{action} of \textbf{surfing}.} \\
\end{tabular}}

\resizebox{\linewidth}{!}{
\begin{tabular}{p{4.8cm}p{4.5cm}p{4.5cm}p{4.5cm}p{4.5cm}}
&&&&\\
\includegraphics[height=3cm]{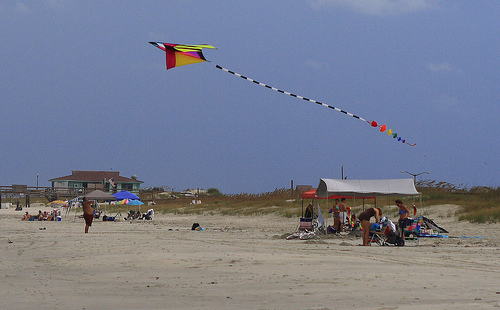}&\includegraphics[height=3cm]{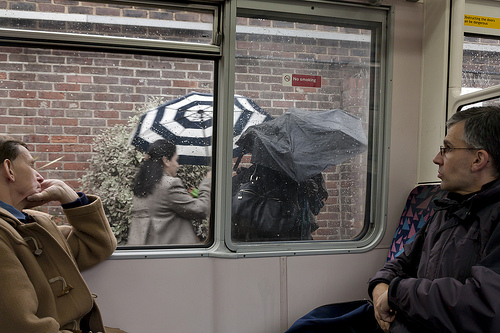} &\includegraphics[height=3cm,width=4.5cm]{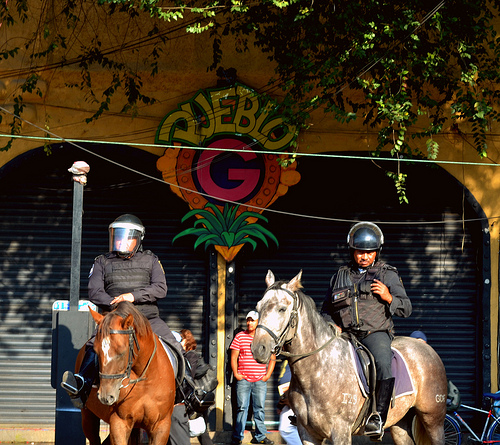} &\includegraphics[height=3cm,width=4.5cm]{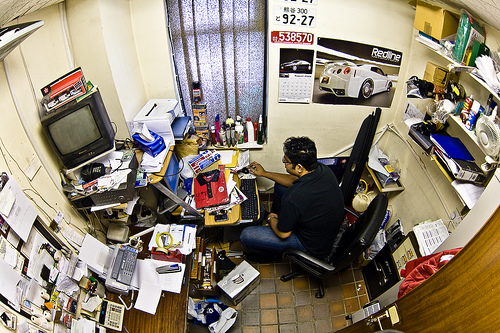} & \includegraphics[height=3cm,width=4.5cm]{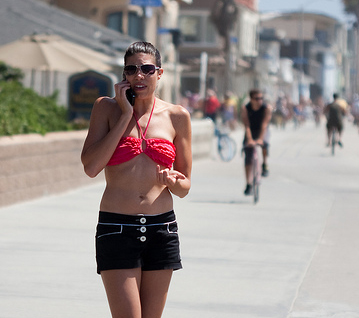} \\ 
\textbf{Q:} {\color{red}{Where}} {\color{cyan}is} the {\color{blue}kite}?&\textbf{Q:} How is {\color{cyan}the} {\color{red}weather} {\color{blue}outside}?&\textbf{Q:} What {\color{blue}the} {\color{cyan}policemen} {\color{red}riding}?&\textbf{Q:} {\color{red}Where} is {\color{cyan}this} {\color{blue}place}?&\textbf{Q:} {\color{red}Who} is talking on {\color{cyan}a} {\color{blue}phone}?\\
\textbf{A:} in the sky&\textbf{A:} rainy&\textbf{A:} horse&\textbf{A:} office&\textbf{A:} woman\\
%&&&&\\
\includegraphics[height=3cm]{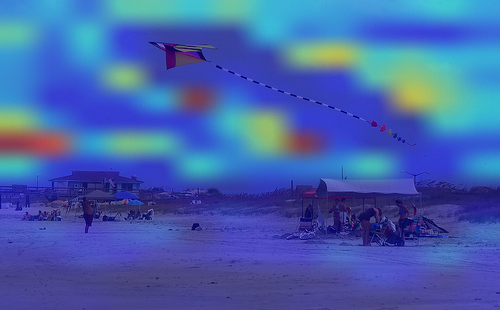}&\includegraphics[height=3cm]{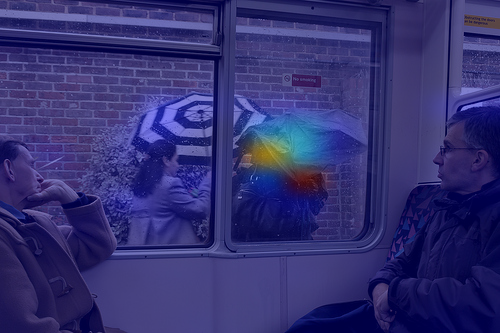} &\includegraphics[height=3cm,width=4.5cm]{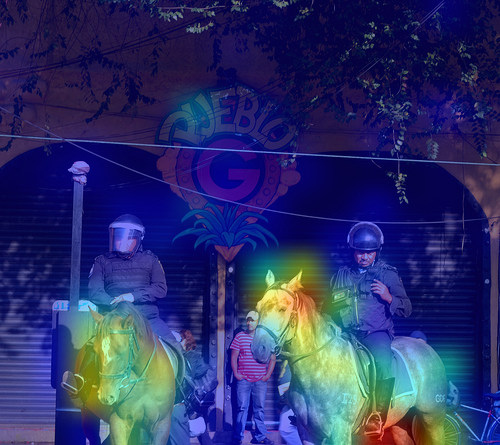} &\includegraphics[height=3cm,width=4.5cm]{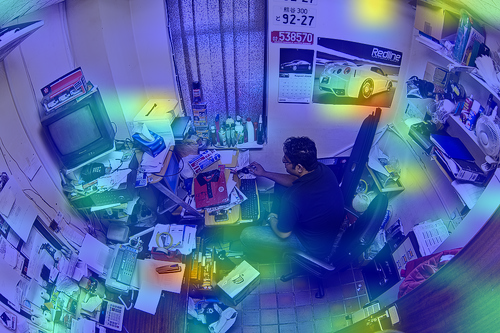} & \includegraphics[height=3cm,width=4.5cm]{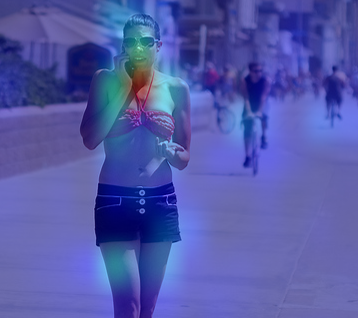} \\
%&&&&\\
\tableincell{p{4.8cm}}{$\bullet$ The \textbf{kite} is \textbf{in} the \textbf{sky}. \\$\bullet$ This image contains the \textbf{object} of \textbf{kite}. \\ $\bullet$ This image happens in the \textbf{scene} of \textbf{beach}.} &\tableincell{p{4.5cm}}{$\bullet$ The \textbf{woman} is \textbf{with} the \textbf{umbrella}.\\ $\bullet$ This image contains the \textbf{attribute} of \textbf{rain}. \\ $\bullet$ This image contains the \textbf{object} of \textbf{umbrella}.}&\tableincell{p{4.4cm}}{$\bullet$ The \textbf{military officer} is \textbf{on} the \textbf{horse}.\\ $\bullet$ The \textbf{military officer} is \textbf{riding} the \textbf{horse}. \\ $\bullet$ This image contains the \textbf{object} of \textbf{police}.}&\tableincell{p{4.4cm}}{$\bullet$ This image contains the \textbf{object} of \textbf{desk}.\\ $\bullet$ The \textbf{printer} is \textbf{on} the \textbf{desk}. \\ $\bullet$ This image happens in the \textbf{scene} of \textbf{office}.}&\tableincell{p{4.4cm}}{$\bullet$ The \textbf{woman} is \textbf{holding} the \textbf{telephone}.\\ $\bullet$ The \textbf{woman} is \textbf{wearing} the \textbf{sunglasses}. \\ $\bullet$ The \textbf{woman} is \textbf{wearing} the \textbf{short pants}.} \\
\end{tabular}}
%\vspace{1pt}

\caption{Some qualitative results produced by our complete model on the Visual Genome QA test split. Image, QA pair, attention map and predicted Top-3 reasons are shown in order. Our model is capable of co-attending between question, image and supporting \facts. The colored words in the question have Top-3 identified weights, ordered as {\color{red}red}, {\color{blue}blue} and {\color{cyan}cyan}. The highlighted area in the attention map indicates the attention weights on the image regions (from red: high to blue: low). The Top-3 identified \facts are re-formulated as human readable reasons, shown as bullets.}
\label{results_examples}
\vspace{-10pt} 
\end{figure*}

\vspace{-0.3cm}
\subsubsection{Evaluation with Predicted \CapFacts}
\vspace{-0.2cm}
In a normal VQA setting the ground truth \facts are not provided, so we now evaluate our model using predicted \facts. All \facts were predicted by models that have been pre-trained on different computer vision tasks, \eg object detection, attributes prediction, relationship prediction and scene classification (see Sec. \ref{imp_details} for more details).

The \textbf{PredFact(Obj+Att+Rel)} model uses all of the predicted \facts as the input, while \textbf{PredFact(Obj+Att+Rel+Extra)} additionally uses the predicted scene category, which is trained on a different data source, the MIT Places 205 \cite{zhou2014learning}. From Table \ref{vg_results}, we see that although all \facts have been included, the performance of these two \facts-only models is still lower than that of the previous state-of-the-art model \textbf{HieCoAtt-VGG}. Considering that errors in \fact prediction may pass to the question answering part and the image features are not used, these results are reasonable. If the \fact prediction models perform better, the final answer accuracy will be higher, as shown in the previous ground-truth \facts experiments. The \textbf{PredFact(Obj+Att+Rel+Extra)} outperforms the \textbf{PredFact(Obj+Att+Rel)} model by 1.2\%, because the extra scene category \facts are included. This suggests that our model benefits by using multiple \helpers, even though they are trained on different data sources, and on different tasks. And as more \facts are added, our model performs better. Compared with the baseline model \textbf{VGG+Obj+Att+Rel+Extra+LSTM}, our \textbf{PredFact(Obj+Att+Rel+Extra)} outperforms it by a large margin, which suggests that our co-attention model can more effectively exploit the \fact-based information.

Image features are further inputted to our model, producing two variants \textbf{PredFact(Obj+Att+Rel)+VGG} and \textbf{PredFact(Obj+Att+Rel+Extra)+VGG}. Both of these outperform the state of art, and our complete model \textbf{PredFact(Obj+Att+Rel+Extra)+VGG} outperforms it by 1\%, \ie more than 6,000 more questions are correctly answered. 

Please note that all of these results are produced by using the naive \facts extraction models described in Sec. \ref{imp_details}. We believe that as better \facts extraction models (such as the relationship prediction models from~\cite{lu2016visual}, for instance) become available, the results will improve further.

\rowcolors{4}{gray!25}{} % Zebra rows (should start on row 4 but does not seem to work: the first gray row appears is on row 3... crappy latex)
\begin{table*}[t!]
\centering
\resizebox{\linewidth}{!}{
\begin{tabular}{llccccccccccccccccccc}
\Xhline{2\arrayrulewidth}
\hiderowcolors
\multicolumn{1}{l}{} & \multicolumn{1}{l}{} & \multicolumn{9}{c}{Test-\textbf{dev}}                                            &  & \multicolumn{9}{c}{Test-\textbf{std}}                                       \\ \cline{3-11} \cline{13-21} 
\multicolumn{2}{c}{Method}                  & \multicolumn{4}{c}{Open-Ended} &  & \multicolumn{4}{c}{Multiple-Choice} &  & \multicolumn{4}{c}{Open-Ended} &  & \multicolumn{4}{c}{Multiple-Choice} \\ \cline{3-6} \cline{8-11} \cline{13-16} \cline{18-21} 
                     &                      & Y/N   & Num   & Other  & All   &  & Y/N    & Num     & Other    & All   &  & Y/N   & Num    & Other  & All  &  & Y/N    & Num     & Other    & All   \\
\Xhline{2\arrayrulewidth}
\showrowcolors
%NMN \cite{andreas2015deep}          &       & 77.7  & 37.2  & 39.3   & 54.8  &  &  -     &  -      &  -       &  -    &  & -     &  -     &  -     & 55.1 &  &  -     &  -      &  -       &  -    \\
iBOWING \cite{zhou2015simple}       &       & 76.6  & 35.0  & 42.6   & 55.7  &  & 76.7   & 37.1    & 54.4     & 61.7  &  & 76.8  & 35.0   & 42.6   & 55.9 &  & 76.9   & 37.3    & 54.6     & 62.0  \\
MCB-VGG \cite{fukui2016multimodal}  &       & -  & -  & -   & 57.1  &  &  -     &  -      &  -       &  -    &  & -     &  -     &  -     & - &  &  -     &  -      &  -       &  -    \\
DPPnet \cite{Noh2015image}          &       & 80.7  & 37.2  & 41.7   & 57.2  &  & 80.8   & 38.9    & 52.2     & 62.5  &  & 80.3  & 36.9   & 42.2   & 57.4 &  & 80.4   & 38.8    & 52.8     & 62.7  \\
D-NMN \cite{andreas2016learning}    &       & 80.5  & 37.4  & 43.1   & 57.9  &  &  -     &  -      &  -       &  -    &  & -     &  -     &  -     & 58.0 &  &  -     &  -      &  -       &  -    \\
VQA team \cite{antol2015vqa}        &       & 80.5  & 36.8  & 43.1   & 57.8  &  & 80.5   & 38.2    & 53.0     & 62.7  &  & 80.6  & 36.4   & 43.7   & 58.2 &  & 80.6   & 37.7    & 53.6     & 63.1  \\
SMem \cite{xu2015ask}               &       & 80.9  & 37.3  & 43.1   & 58.0  &  &  -     &  -      &  -       &  -    &  & 80.8  & 37.3   & 43.1   & 58.2 &  &  -     &  -      &  -       &  -    \\
SAN \cite{yang2015stacked}          &       & 79.3  & 36.6  & 46.1   & 58.7  &  &  -     &  -      &  -       &  -    &  &  -    &  -     &  -     & 58.9 &  &  -     &  -      &  -       &  -    \\
ACK \cite{wu2015ask}                &       & 81.0  & 38.4  & 45.2   & 59.2  &  &  -     &  -      &  -       &  -    &  & 81.1  & 37.1   & 45.8   & 59.4 &  &  -     &  -      &  -       &  -    \\
DMN+ \cite{xiong2016}               &       & 80.5  & 36.8  & 48.3   & 60.3  &  &  -     &  -      &  -       &  -    &  &  -    &  -     &  -     & 60.4 &  &  -     &  -      &  -       &  -    \\
MRN-VGG \cite{kim2016multimodal} &       & 82.5  & 38.3  & 46.8   & 60.5  &  &  82.6  & 39.9   &  55.2   & 64.8   &  & -  & -   & -   & - &  & -   & -    & -     & -  \\
HieCoAtt-VGG \cite{lu2016hierarchical}  &       & 79.6  & 38.4  & 49.1   & 60.5  &  & 79.7   &  40.1   & 57.6     & 64.9  &  &  -    &  -     &  -     &  -   &  &  -     &  -      &  -       &  -    \\ \hline
Re-Ask-ResNet \cite{malinowski2016ask} &      & 78.4  & 36.4  & 46.3   & 58.4  &  &  -     &  -      &  -       &  -    &  & 78.2  & 36.3   & 46.3   & 58.4 &  &  -     &  -      &  -       &  -    \\
FDA-ResNet \cite{ilievski2016focused}      &       & 81.1  & 36.2  & 45.8   & 59.2  &  &  -     &  -      &  -       &  -    &  &  -    &  -     &  -     & 59.5 &  &  -     &  -      &  -       &  -    \\
MRN-ResNet \cite{kim2016multimodal} &       & 82.4  & 38.4  & 49.3   & 61.5  &  &  82.4  & 39.7   &  57.2   & 65.6   &  & 82.4  & 38.2   & 49.4   & 61.8 &  & 82.4   & 39.6    & 58.4     & 66.3  \\
HieCoAtt-ResNet \cite{lu2016hierarchical} & & 79.7  & 38.7  & 51.7   & 61.8  &  & 79.7   &  40.0   & 59.8     & 65.8  &  &  -    &  -     &  -     & 62.1 &  &  -     &  -      &  -       & 66.1  \\
MCB-Att-ResNet \cite{fukui2016multimodal}& & 82.5  & 37.6  & 55.6   & 64.7  &  &  -     &  -      &  -       & 69.1  &  &  -    &  -     &  -     &  -   &  &  -     &  -      &  -       &  -    \\ \hline
Ours-VGG                            &       & 81.2 & 37.7  &  50.5   & 61.7   &  &  81.3 &   39.9   &  60.5       & 66.8    &  &  81.3    &  36.7     &  50.9     & 61.9&  &   81.4    &   39.0     &   60.8      &  67.0    \\
Ours-ResNet                         &       & 81.5 & 38.4   & 53.0   & 63.1   &  &  81.5 &   40.0   &  62.2       & 67.7    &  &  81.4    &  38.2     &  53.2     & 63.3&  &   81.4    &    39.8    &  62.3       &  67.8    \\
\Xhline{2\arrayrulewidth}
\end{tabular}}
\caption{Single model performance on the VQA-real test set in the open-ended and multiple-choice settings.}
\label{tab:VQA_result}
\vspace{-8pt}
\end{table*}

\vspace{-0.3cm}
\subsubsection{Human agreements on Predicted Reasons}
\vspace{-0.1cm}
\label{sec:human}
A key differentiator of our proposed model is that 
the weights resulting from the \facts attention process can be used to generate human-interpretable reasons for the answer generated.
To evaluate the human agreement with these generated reasons, we sampled 1,000 questions that have been correctly answered by our \textbf{PredFact(Obj+Att+Rel+Extra)+VGG} model, and conduct a human agreement study on the generated reasons. 

Since this is the first VQA model that can generate human readable reasons, there is no previous work that we can follow to perform the human evaluation. We have thus designed the following human agreement experimental protocols. At first, an image with the question and our correctly generated answer are given to a human subject. Then a list of human readable reasons (ranging from 20 to 40) are shown. These reasons are formulated from \facts that are predicted from the \facts extraction model introduced in the Sec.\ref{imp_details}. Although these reasons are all related to the image, not all of them are relevant to the particular question asked. The task of the human subjects is thus to choose the reasons that are related to answering the question. The human agreements are calculated by matching the human selected reasons with our model ranked reasons. In order to ease the human subjects' workload and to have an unique guide for them to select the `reasons', we ask them to select only the top-1 reason that is related to the question answering. And they can choose nothing if they think none of the provided reasons are useful.

Finally, in the evaluation, we calculate the rate at which the human selected reason can be found in our generated top-1/3/5 reasons. We find that \textbf{30.1\%} of the human selected top-1 reason can be matched with our model ranked top-1 reason. For the top-3 and top-5, the matching rate are \textbf{54.2\%} and \textbf{70.9\%}, respectively. This suggests that 
the reasons generated are both interpretable and informative.
Figure \ref{results_examples} shows some example reasons generated by our model.

\vspace{-1pt}
\subsection{Results on the VQA-real}
\vspace{-1pt}
Table \ref{tab:VQA_result} compares our approach with state-of-the-art on the VQA-real dataset. Since we do not use any ensemble models, we only compare with the single models on the VQA test leader-board. The \textit{test-dev} is normally used for validation while the \textit{test-standard} is the default test data. The first section of Table~\ref{tab:VQA_result} shows the state of art methods that use VGG features, except iBOWING~\cite{zhou2015simple}, which uses the GoogLeNet features~\cite{szegedy2014googlenet}. The second section gives the results of models that use ResNet~\cite{he2015resnet} features. Two versions of our complete model are evaluated at the last section, using VGG and ResNet features, respectively.

\textbf{Ours-VGG} produces the best result on all of the splits, compared with models using the same VGG image encoding method. 
\textbf{Ours-ResNet} ranks the second amongst the single models using ResNet features on the test-dev split, but we achieve the state of the art results on the test-std, for both Open-Ended and Multiple-Choice questions. The best result on the test-dev with ResNet features is achieved by the Multimodal Compact Milinear (MCB) pooling model with the visual attention~\cite{fukui2016multimodal}. We believe the MCB can be integrated within our proposed co-attention model, by replacing the linear embedding steps in Eqs.~\ref{eq:seq_attend} and~\ref{eq:mlp}, but we leave it as a future work.

\vspace{-6pt}
\section{Conclusion}
\vspace{-3pt}
We have proposed a new approach which is capable of adaptively combining the outputs from other algorithms in order to solve a new, more complex problem.  We have shown that the approach can be applied to the problem of Visual Question Answering, and that in doing so it achieves state of the art results. Visual Question Answering is a particularly interesting application of the approach, as in this case the new problem to be solved is not completely specified until run time. In retrospect, it seems strange to attempt to answer general questions about images without first providing access to readily available image information that might assist in the process. In developing our approach we proposed a co-attention method applicable to questions, image and \facts jointly. We also showed that attention-weighted \facts serve to illuminate why the method reached its conclusion, which is critical if such techniques are to be used in practice.  

{\small
\bibliographystyle{ieee}
\bibliography{myref}
}

\appendix

%%%%%%%%% BODY TEXT
\section{Reformatting \CapFacts to Reasons}
In this section we provide additional detail on the method used to reformulate the ranked \facts into human-readable reasons using a template-based approach.

Each of the extracted \facts has been encoded by a structural triplet (see Sec.3.1 for more details) {\ttfamily (subject,}{\ttfamily relation,}{\ttfamily object)}, where {\ttfamily subject} and {\ttfamily object} denote two visual concepts and {\ttfamily relation} represents a relationship between these two concepts. This structured representation enables us to reformulate the \fact to a human-readable sentence easily, with a pre-defined template. All of the templates are shown in Table \ref{tab:triplet}.
	
\begin{table}[h!]
%\vspace{-5pt}
	\begin{center}
		\scriptsize
		\resizebox{\linewidth}{!}{
			\rowcolors{2}{gray!25}{} % Zebra rows (starting on row 2)
			\begin{tabular}{l l }
				\Xhline{2\arrayrulewidth}
									Fact Triplet													&Reason Template \\ \hline
				{\ttfamily (\_img,}{\ttfamily \_scene,}{\ttfamily img\_scn)}		&This image happens in the scene of {\ttfamily img\_scn}.		 \\
				{\ttfamily (\_img,}{\ttfamily \_att,}{\ttfamily img\_att)}	& This image contains the attribute of {\ttfamily img\_att}.		 \\
				{\ttfamily (\_img,}{\ttfamily \_contain,}{\ttfamily obj)}			& This image contains the object of {\ttfamily obj}.	\\ 
				{\ttfamily (obj,}{\ttfamily \_att,}{\ttfamily obj\_att)}		& The {\ttfamily obj} is {\ttfamily obj\_att}.												\\
				{\ttfamily (obj1,}{\ttfamily rel,}{\ttfamily obj2)}					& The {\ttfamily obj1} is {\ttfamily rel} the {\ttfamily obj2}.	\\
				\Xhline{2\arrayrulewidth}
			\end{tabular}}
			\vspace{-1pt}
			\caption{\CapFacts represented by triplets and the corresponding reason template. 
				{\ttfamily \_img}, {\ttfamily \_scene}, {\ttfamily \_att} and {\ttfamily \_contain} are specific tokens. 
				While {\ttfamily img\_scn}, {\ttfamily obj}, {\ttfamily img\_att}, {\ttfamily obj\_att} and {\ttfamily rel} refer to 
							vocabularies describing image scenes, objects, image/object attributes and relationships between objects. These words are filled into the pre-defined template to generate reasons.
				} 
			\label{tab:triplet}
		\end{center}
		\vspace{-20pt}
	\end{table}	
	
\begin{table}[h!]
%\vspace{-5pt}
	\begin{center}
		\scriptsize
		\resizebox{\linewidth}{!}{
			\rowcolors{2}{gray!25}{} % Zebra rows (starting on row 2)
			\begin{tabular}{l l }
				\Xhline{2\arrayrulewidth}
									Example Fact													&Example Reason \\ \hline
				{\ttfamily (\_img,}{\ttfamily \_scene,}{\ttfamily \textbf{office})}&This image happens in the scene of {\ttfamily \textbf{office}}.		 \\
				{\ttfamily (\_img,}{\ttfamily \_att,}{\ttfamily \textbf{wedding})}&This image contains the attribute of {\ttfamily \textbf{wedding}}.		 \\
				{\ttfamily (\_img,}{\ttfamily \_contain,}{\ttfamily \textbf{dog})}&This image contains the object of{\ttfamily \textbf{dog}}.	\\ 
				{\ttfamily (\textbf{shirt},}{\ttfamily \_att,}{\ttfamily \textbf{red})}&The {\ttfamily \textbf{shirt}} is {\ttfamily \textbf{red}}.												\\
				{\ttfamily (\textbf{man},}{\ttfamily \textbf{hold},}{\ttfamily \textbf{umbrella})}&The {\ttfamily \textbf{man}} is {\ttfamily \textbf{hold}} the {\ttfamily \textbf{umbrella}}.	\\
				\Xhline{2\arrayrulewidth}
			\end{tabular}}
			\vspace{-1pt}
			\caption{Example \facts in triplet and their corresponding generated reasons based on the templates in the previous table.
				} 
			\label{tab:example}
		\end{center}
		\vspace{-20pt}
	\end{table}

For the {\ttfamily (\_img,}{\ttfamily \_att,}{\ttfamily img\_att)} triplet, since we also have the super-class label of the {\ttfamily img\_att} vocabulary, such as `action', `number', `attribute' \etc, we can generate more varieties.
%A*** The end of that last sentence needs replacing
For example, we can generate a reason such as `This image contains the \textit{action} of {\ttfamily surfing}', because `surfing' belongs to the super-class of `action'.

\section{WUPS Evaluation on the Visual Genome QA}
The WUPS calculates the similarity between two words based on the similarity between their common subsequence in the taxonomy tree. If the similarity between two words is greater than a threshold then the candidate answer is considered to be right. We report on thresholds 0.9 and 0.0, following~\cite{ma2015learning,ren2015image}. Table \ref{vg_results_WUPS9} and \ref{vg_results_WUPS0} show the question type-specific results.

\begin{table}[h]
\centering
\resizebox{\linewidth}{!}{
%\rowcolors{1}{gray!25}{} % Zebra rows (starting on row 2)
\begin{tabular}{lccccccc}
\Xhline{2\arrayrulewidth}
\multicolumn{1}{c}{}                                    & \multicolumn{7}{c}{WUPS @ 0.9 (\%)}                                      \\  \Cline{0.8pt}{2-8} 
\multicolumn{1}{c}{Methods}                             & What     & Where    & When    & Who     & Why     & How      & \multirow{2}{*}{\textbf{Overall}} \\
\multicolumn{1}{c}{}                                    & (60.5\%) & (17.0\%) & (3.5\%) & (5.5\%) & (2.7\%) & (10.8\%) & \\ \Xhline{2\arrayrulewidth}
\rowcolor{gray!30}VGG+LSTM \cite{antol2015vqa}          & 42.84    & 18.61    & 55.81   & 35.75   & 12.53   & 45.78    & 38.30\\
VGG+Obj+Att+Rel+Extra+LSTM                              & 44.44    & 19.12    & 55.85   & 37.94   & 12.62   & 46.98    & 39.61\\ 
\rowcolor{gray!30}HieCoAtt-VGG \cite{lu2016hierarchical}& 47.38    & 19.89    & 55.83   & 39.40   & 13.72   & 48.15    & 41.75\\ \Xhline{1.5\arrayrulewidth}
GtFact(Obj)                                             & 45.86    & 20.05    & 55.19   & 42.99   & 13.86   & 48.80    & 40.83\\
\rowcolor{gray!25}GtFact(Obj+Att)                       & 49.74    & 19.87    & 55.53   & 42.95   & 13.98   & 47.00    & 43.24\\
GtFact(Obj+Rel)                                         & 46.35    & 20.45    & 54.93   & 43.75   & 13.89   & 46.42    & 41.25\\
\rowcolor{gray!25}GtFact(Obj+Att+Rel)                   & 50.45    & 20.55    & 54.98   & 43.81   & 14.15   & 47.37    & 43.86\\
GtFact(Obj+Att+Rel)+VGG                                 & 51.69    & 21.18    & 55.90   & 44.06   & 14.05   & 49.03    & 44.94\\\Xhline{1.5\arrayrulewidth}
\rowcolor{gray!20}PredFact(Obj+Att+Rel)                 & 44.87    & 19.46    & 54.97   & 39.70   & 13.89   & 46.04    & 39.92\\
PredFact(Obj+Att+Rel+Extra)                             & 46.23    & 20.27    & 55.30   & 40.31   & 13.95   & 47.37    & 41.08\\
\rowcolor{gray!20}PredFact(Obj+Att+Rel)+VGG             & 47.86    & 20.17    & 55.77   & 40.35   & 13.81   & 48.34    & 42.16\\
PredFact(Obj+Att+Rel+Extra)+VGG                         & 48.48    & 20.69    & 55.76   & 40.96   & 13.97   & 48.97    & 42.73\\\Xhline{2\arrayrulewidth}
\end{tabular}}
\caption{Ablation study on the Visual Genome QA dataset. WUPS at 0.9 for different question types are shown. The percentage of questions for each type is shown in parentheses.}
\vspace{-10pt}
\label{vg_results_WUPS9}
\end{table}

\begin{table}[h]
\centering
\resizebox{\linewidth}{!}{
%\rowcolors{1}{gray!25}{} % Zebra rows (starting on row 2)
\begin{tabular}{lccccccc}
\Xhline{2\arrayrulewidth}
\multicolumn{1}{c}{}                                    & \multicolumn{7}{c}{WUPS @ 0.0 (\%)}                                      \\  \Cline{0.8pt}{2-8} 
\multicolumn{1}{c}{Methods}                             & What     & Where    & When    & Who     & Why     & How      & \multirow{2}{*}{\textbf{Overall}} \\
\multicolumn{1}{c}{}                                    & (60.5\%) & (17.0\%) & (3.5\%) & (5.5\%) & (2.7\%) & (10.8\%) & \\ \Xhline{2\arrayrulewidth}
\rowcolor{gray!30}VGG+LSTM \cite{antol2015vqa}          & 66.49    & 27.39    & 63.29   & 55.25   & 15.99   & 72.25    & 58.39\\
VGG+Obj+Att+Rel+Extra+LSTM                              & 67.10    & 27.64    & 63.25   & 56.05   & 16.12   & 72.61    & 58.89\\ 
\rowcolor{gray!30}HieCoAtt-VGG \cite{lu2016hierarchical}& 68.41    & 28.50    & 62.97   & 56.83   & 17.41   & 73.33    & 59.97\\ \Xhline{1.5\arrayrulewidth}
GtFact(Obj)                                             & 67.94    & 28.57    & 62.27   & 58.19   & 17.42   & 72.78    & 59.69\\
\rowcolor{gray!25}GtFact(Obj+Att)                       & 69.07    & 28.37    & 62.68   & 58.35   & 17.62   & 73.06    & 60.39\\
GtFact(Obj+Rel)                                         & 68.22    & 28.85    & 62.10   & 58.60   & 17.51   & 72.66    & 59.91\\
\rowcolor{gray!25}GtFact(Obj+Att+Rel)                   & 69.43    & 28.85    & 62.10   & 58.67   & 18.00   & 73.24    & 60.72\\
GtFact(Obj+Att+Rel)+VGG                                 & 69.99    & 29.29    & 62.91   & 58.56   & 17.70   & 73.80    & 61.21\\\Xhline{1.5\arrayrulewidth}
\rowcolor{gray!20}PredFact(Obj+Att+Rel)                 & 67.36    & 28.27    & 62.30   & 56.73   & 17.54   & 72.68    & 59.20\\
PredFact(Obj+Att+Rel+Extra)                             & 67.98    & 28.91    & 62.56   & 57.11   & 17.70   & 72.96    & 59.75\\
\rowcolor{gray!20}PredFact(Obj+Att+Rel)+VGG             & 68.55    & 28.71    & 62.70   & 57.01   & 17.43   & 73.32    & 60.09\\
PredFact(Obj+Att+Rel+Extra)+VGG                         & 68.88    & 29.07    & 62.91   & 57.28   & 17.76   & 73.38    & 60.39\\\Xhline{2\arrayrulewidth}
\end{tabular}}
\caption{Ablation study on the Visual Genome QA dataset. WUPS at 0.0 for different question types are shown. The percentage of questions for each type is shown in parentheses.}
\vspace{-10pt}
\label{vg_results_WUPS0}
\end{table}

\end{document}